\documentclass{article}
\usepackage{spconf,amsmath,graphicx}
\usepackage{subfigure} 

\usepackage{algorithm}
\usepackage{algorithmic}
\usepackage{nohyperref} 

\usepackage{amsmath,amsfonts,amssymb,amsthm}
\usepackage{bbm,bm}

\usepackage{multirow}
\usepackage{booktabs}

\title{Sensing-Aware Kernel SVM}
\name{Weicong Ding, Prakash Ishwar, Venkatesh Saligrama, and W. Clem Karl \thanks{This article is based upon work supported by the U.S. AFOSR and the U.S. NSF under award numbers $\sharp$FA9550-10-1-0458 (subaward $\sharp$A1795) and $\sharp$1218992 respectively.}}
\address{Department of Electrical and Computer Engineering, Boston
  University, Boston, MA, USA. \\ \{{\tt dingwc,pi,srv,wckarl}\}{\tt @bu.edu}}

\begin{document}
%
\maketitle
\begin{abstract}
We propose a novel approach for designing kernels for support vector
machines (SVMs) when the class label is linked to the observation
through a latent state and the likelihood function of the observation
given the state (the sensing model) is available. We show that the
Bayes-optimum decision boundary is a hyperplane under a mapping
defined by the likelihood function. Combining this with the maximum
margin principle yields kernels for SVMs that leverage knowledge of
the sensing model in an optimal way. We derive the optimum kernel for
the bag-of-words (BoWs) sensing model and demonstrate its superior
performance over other kernels in document and image classification
tasks. These results indicate that such optimum sensing-aware kernel
SVMs can match the performance of rather sophisticated
state-of-the-art approaches.
\end{abstract}
\begin{keywords}
Sensing model, Kernel method, SVM, Bag of Words, Supervised Classification
\end{keywords}

\vspace*{-1ex}
\section{Introduction}
\vspace*{-1ex}
This paper presents a new provably optimum method for designing
kernels for SVMs that explicitly incorporates information about the
structure of the underlying data generating process.
We consider the typical classification task where the observed data
point $\mathbf{x}\in\mathcal{X}$ and label $y$ follow some joint
distribution $p(\mathbf{x}, y)$.
In many real world problems, however, observations $\mathbf{x}$ are
only indirectly related through some generative process to latent
variables $\mathbf{z}\in\mathcal{Z}$ via a {\it sensing} model
$p(\mathbf{x}|\mathbf{z})$ and the latent variables have different
distributions conditioned on the underlying label $y$. 
An example of such a situation is medical tomography, where the
acquired data $\mathbf{x}$ consists of X-ray based projections but the
diagnosis is done on reconstructed cross-sections of the body
$\mathbf{z}$ to determine the presence or absence of a disease $y$.
Another example is the classic ``Bag-of-Word'' (BoW) modeling paradigm
widely-used to model text documents, images,
etc.~\cite{Blei2012Review:ref}\cite{Li05:ref}, where the observed
document $\mathbf{x}$ is modeled as a collection of iid drawings of
words from a latent document-specific distribution $\mathbf{z}$ over
the vocabulary. This $\mathbf{z}$ is, in turn, related to the document
category $y$ through an unknown $p(\mathbf{z}|y)$.

For supervised classification, a generative approach would make
further assumptions on $p(\mathbf{z}\vert y)$ which may either not
hold or lead to an intractable posterior inference
problem~\cite{DiscLDA08:ref}.
On the other hand, the classical distribution-free discriminative
paradigm would ignore knowledge of the sensing model
$p(\mathbf{x}|\mathbf{z})$ and build classifiers using labeled
training data~\cite{Vapnik:ref,hastie2009elements}.
%
%
%
%
Yet another approach is a decoupled two-step process where the latent
states are first estimated as $\hat{\mathbf{z}}(\mathbf{x})$ using the
sensing model and then a classifier is built based on
$(\hat{\mathbf{z}},y)$ pairs. Computing estimates, however, can be
costly and can introduce confounding artifacts when data is limited or
noisy, and may be unnecessary if the ultimate goal is a simple
decision.
%
%

%
%
Motivated by such problems, we start by examining the structure
of the Bayes-optimum classifier, focusing on binary classification,
and then connect it to kernel SVMs via the max-margin principle. Our proposed
``sensing-aware'' kernel-design arises as the natural consequence of
such a procedure.  We illustrate the approach on the BoW model and
demonstrate its merits on document and image classification tasks.

\vspace*{-2ex}
\section{Optimum Sensing-Aware Kernel}
\vspace*{-1ex}
\label{sec:sensing}
Let $y \in \{-1,+1\}$, $\mathbf{z} \in \mathcal{Z}$, and $\mathbf{x}
\in \mathcal{X}$ denote the class label, latent variable, and
observation respectively.
$\mathbf{x}$ and $y$ are independent conditioned on
$\mathbf{z}$ and the {\it sensing model} $p(\mathbf{x}\vert
\mathbf{z})$ is known. 
We make the following technical assumptions on the joint distribution
which are satisfied in many applications.

\noindent {\bf Assumption 1:} {\it (Square-integrability)} $\forall
\mathbf{x}, y, p(\mathbf{x}\vert \mathbf{z}), p(\mathbf{z},y) \in
\mathcal{L}^{2}(\mathcal{Z})$ as functions of $\mathbf{z}$.

\noindent {\bf Assumption 2:} {\it (Uniform boundedness)} There exists
$R \in [0,\infty)$ such that $\sup_{x,z}\vert p(\mathbf{x}\vert
  \mathbf{z}) \vert \leq R$.

\noindent Further, let $\langle f, g\rangle_{\mathcal{Z}} := \int_{\mathcal{Z}}
f(\mathbf{z})g(\mathbf{z})d\mathbf{z}$ denote the usual inner product
in $\mathcal{L}^{2}(\mathcal{Z})$ which induces the norm $\Vert
f\Vert_{\mathcal{Z}}^2 \triangleq \langle f, f
\rangle_{\mathcal{Z}}$. Let $\Vert f \Vert_1 \triangleq
\int_{\mathcal{Z}}\vert f(\mathbf{z})\vert d\mathbf{z}$ denote the
$\ell_1$-norm in $\mathcal{L}^{2}(\mathcal{Z})$.

When full knowledge of the generative model $p(\mathbf{x}, y)$ is
available, it is well known that the {\it Maximum Aposteriori
  Probability} (MAP) estimator of the class label minimizes the error
probability, i.e., the Bayes risk. For binary classification, this
reduces to a simple likelihood ratio test (LRT):
%
\begin{equation}
\label{eqa:lrt}
p(y=1\vert\mathbf{x}) \overset{1}{\underset{-1}{\gtrless}} p(y=-1\vert
\mathbf{x})
\end{equation}
%
%
%
Using $p(\mathbf{x},\mathbf{z},y) = p(\mathbf{x}|\mathbf{z})
p(\mathbf{z},y)$ and marginalizing over $\mathbf{z}$ the LRT
\eqref{eqa:lrt} reduces to,
\vspace*{-2ex}
\begin{equation}
\label{eqa:sensing-principle}
\langle p(\mathbf{x}\vert \mathbf{z}), ~w(\mathbf{z})
\rangle_{\mathcal{Z}} \overset{1}{\underset{-1}{\gtrless}} 0
\end{equation}
\vglue -2ex
\noindent where $w(\mathbf{z}) := p(\mathbf{z},y=1) -
p(\mathbf{z},y=-1)$. Note that $w (\mathbf{z})$ satisfies the
following constraint
\vspace*{-2ex}
\begin{equation*}
\label{wz:constraint}
\Vert w(\mathbf{z}) \Vert_1 \leq \Vert p(\mathbf{z},y=1) \Vert_1 +
\Vert p(\mathbf{z}, y=-1) \Vert_1 = 1.
\end{equation*}
\vglue -2ex

%
\noindent {\bf Theorem:} If we interpret $p(\mathbf{x}\vert
\mathbf{z})$ and $w(\mathbf{z})$ as vectors (functions of ${\mathbf
  z}$ with ${\mathbf x}$ held fixed) in
$\mathcal{L}^{2}(\mathcal{Z})$, then the Bayes-optimal rule
\eqref{eqa:sensing-principle} is a {\it linear} classifier in
$\mathcal{L}^{2}(\mathcal{Z})$ defined by the separating hyperplane
$w({\mathbf{z}})$.
When a set of training samples $\{\mathbf{x}^i,y^i\}_{i=1}^{M}$ that
are generated from $p(\mathbf{x},y)$ in an iid fashion are given, and
$w(\mathbf{z})$ is unknown, the popular {\bf max-margin principle} advocates
using the hyperplane $w^{*}(\mathbf{z})$ that maximizes the
separation, i.e., a (soft) margin, between the two classes. This
max-margin hyperplane $w^{*}(\mathbf{z})$ is the optimal solution of
the following (infinite-dimensional) constrained optimization problem
\vspace*{-1ex}
\begin{equation}
\label{eqa:max-margin}
 \begin{aligned}
 w^{*}(\mathbf{z}) = & \underset{w(\mathbf{z})\in \mathcal{L}^{2}(\mathcal{Z}),
     \xi_i}{\arg\min} & & \frac{1}{2}\Vert w(\mathbf{z})
   \Vert_{\mathcal{Z}}^{2} + C\sum\limits_{i=1}^{N}\xi_i \\
 & \text{such that} & & y^i \langle p(\mathbf{x}^i \vert \mathbf{z}),
   ~ w(\mathbf{z})\rangle \geq 1-\xi_i, ~~ \xi_i \geq 0
   \\ 
 & & & \Vert w(\mathbf{z}) \Vert_1\leq 1
 \end{aligned}
\end{equation}
\vglue -1ex
\noindent And the optimum classifier is then given by 
\vspace*{-1ex}
\[
f^{*}(\mathbf{x}) = \mathrm{sign} \left( \langle p(\mathbf{x} \vert
\mathbf{z}), ~ w^{*}(\mathbf{z})\rangle \right).
\]
\vglue -1ex
\noindent If we write $w^{*}(\mathbf{z}) = \sum\limits_{i=1}^{M} \beta_i
p(\mathbf{x}^i|\mathbf{z}) + w^{\perp}(\mathbf{z})$, where $w^{\perp}$
is orthogonal to the linear span of $\{ p(\mathbf{x}_i|\mathbf{z}) \}_{i=1}^{M}$, and ignore the constraint $\Vert w(\mathbf{z}) \Vert
\leq 1$, the optimization problem \eqref{eqa:max-margin} reduces to
\vspace*{-2ex}
\begin{equation}
\label{eqa:kernel-SVM}
 \begin{aligned}
 & \underset{\beta_i,
     \xi_i, w^{\perp}}{\min} & & \frac{1}{2}\sum\limits_{i,j=1}^{M}\beta_i\beta_j K(\mathbf{x}^i,\mathbf{x}^j) + \Vert w^{\perp} \Vert_{2}^{2} + C\sum\limits_{i=1}^{N}\xi_i  \\
 & \text{such that} & & y^i \sum\limits_{j=1}^{M}\beta_j K(\mathbf{x}^i,\mathbf{x}^j) \geq 1-\xi_i, ~~ \xi_i \geq 0
   \\ 
 \end{aligned}
\end{equation}
\noindent where
\vspace*{-2ex}
\begin{equation}
\label{eqa:kerneldef}
K(\mathbf{x},\mathbf{x}^{\prime}) \triangleq \langle p(\mathbf{x}\vert
\mathbf{z}), ~ p(\mathbf{x}^{\prime}\vert \mathbf{z})
\rangle = \int_{\mathcal{Z}} p(\mathbf{x}\vert
\mathbf{z}) p(\mathbf{x}^{\prime}\vert \mathbf{z}) d{\mathbf{z}}.
\end{equation}
\vglue -2ex
Note that the optimal solution of \eqref{eqa:kernel-SVM} has
$w^{\perp} = 0$ making it a finite-dimensional constrained
optimization problem. Equation \eqref{eqa:kernel-SVM} is, in fact, a
kernel SVM problem with the kernel defined by equation
\eqref{eqa:kerneldef}. We will refer to \eqref{eqa:kerneldef} as the
``sensing-aware kernel'' in what follows.
The optimal classifier $f^{*}(\mathbf{x})$ corresponding to the
solution to \ref{eqa:kernel-SVM} has the following form
\vspace*{-1ex}
\begin{equation}
f(\mathbf{x}) = \mathrm{sign}\left(\sum\limits_{i=1}^{n}\beta_i^{*}
K(\mathbf{x}, \mathbf{x}^i)\right).
\end{equation}
\vglue -1ex

The sensing-aware kernel thus provides a {\it principled} way to
incorporate the partial information about the generative model
$p(\mathbf{x}\vert \mathbf{z})$.
For a wide range of $p(\mathbf{x}\vert \mathbf{z})$,
$K(\mathbf{x},\mathbf{x}^{\prime})$ has a closed form expression and
can be calculated efficiently.
In this paper, we focus on the BoW generative model.

\vspace*{1ex}
\noindent {\bf Remark 1:} 
%
%
If $\mathbf{t}(\mathbf{x})$ is a sufficient statistic for
$\mathbf{z}$, i.e., $\mathbf{x} - \mathbf{t}(\mathbf{x}) - \mathbf{z}$
is a Markov chain, then $\mathbf{t}(\mathbf{x})$ can replace
$\mathbf{x}$ in the above analysis and the resulting kernel SVM will
have the kernel $K(\mathbf{x},\mathbf{x}^{\prime}) =
\int_{\mathcal{Z}} p(\mathbf{t}(\mathbf{x})\vert \mathbf{z})
p(\mathbf{t}(\mathbf{x}^{\prime})\vert \mathbf{z}) d{\mathbf{z}}$.

%
%
\noindent {\bf Remark 2:} There is a vast literature dedicated to
kernel methods for classification\cite{hastie2009elements}. While the
standard RBF kernel works well in many problems, various kernels have
been designed for specific tasks. Among them, a line of work on
model-based kernels is related to our approach in the sense that
knowledge of a generative model is utilized.
The Probability Product Kernel (PPK) proposed in \cite{Jebara04:ref}
first estimates latent variable $\hat{\mathbf{z}}^{i}$ using
observation $\mathbf{x}^{i}$ and defines the kernel in terms of the
estimate as $K(\mathbf{x}^{i},\mathbf{x}^{j}) :=
\int_{\mathcal{X}}p(\mathbf{x}|\hat{\mathbf{z}}^{i})p(\mathbf{x}|\hat{\mathbf{z}}^{j})d\mathbf{x}$.
Heat or Diffusion kernels (Diff) that exploit the Fisher information
metric on the probability manifold were proposed in
\cite{Lafferty05:ref}. Kernels based on the KL divergence were
proposed in \cite{KLdiv:ref}.
Although each approach has its own strength, unlike ours, the
aforementioned probabilistic kernels are not designed to directly
minimize the classification error. 
Moreover, they require full model knowledge which is unreasonable for
large and complex datasets like text or images.

\vspace*{-1ex}
\section{Kernel for Bag of Words models}
\label{sec:bow}
\vspace*{-1ex}
In the ``Bag of words'' (BoW), or ``Bag of Features'' (BoF) model,
there is a collection of {\it documents} composed of words from a
vocabulary of size $W$. Each document is modeled as being generated by
$N$ i.i.d drawings of words from a latent $W\times 1$ document
word-distribution vector $\mathbf{z}$.  Since the ordering is ignored,
a document is represented as a $W\times 1$ empirical word-count vector
$\mathbf{x} = (x_1,\ldots, x_W)^T$. Therefore,
$p(\mathbf{x}\vert\mathbf{z})$ is a multinomial distribution
\vspace*{-2ex}
\begin{equation*}
p(\mathbf{x}\vert\mathbf{z}) = \binom {N} {x_1,\ldots,
  x_W}\prod\limits_{w=1}^{W}z_{w}^{x_w}
\end{equation*}
\vglue -1ex
\noindent From this it can be shown that the sensing kernel for two documents
$\mathbf{x}^{i}, \mathbf{x}^{j}$ defined in \ref{eqa:kerneldef} has
the following form,
\vspace*{-1ex}
\begin{equation}
\label{eqa:bowkernel}
\begin{split}
K(\mathbf{x}^{i}, \mathbf{x}^{j}) &= \prod\limits_{w=1}^{W}
\frac{\Gamma( x_{w}^{i}+x_{w}^{j}+1)}{\Gamma ( x_{w}^{i} +1) \Gamma(
  x_{w}^{j} +1) } \frac{\Gamma(N_i+1)\Gamma(N_j+1)}{\Gamma(N_i+N_j+W)}
\\ 
& = \prod\limits_{w=1}^{W}
\frac{(x_{w}^{i}+x_{w}^{j})!}{(x_{w}^{i})!(x_{w}^{j})!}
\frac{N_i!N_j!}{(N_i+N_j+W-1)!}
\end{split}
\end{equation}
\vglue -2ex
\noindent where $N_i$ is the number of words in the $i$-th document and
$\Gamma(t)$ is the ordinary Gamma function. 
One practical concern is the size of the factorials. Typically, $W$ is
a large constant and $N_i$ varies across documents. Therefore, the
product of factorials will have a large dynamic range and lead to
overflow errors (even for small $W$). The product is sensitive to the
differences in the $N_i$ values across documents.
These issues can be addressed by using one of the following
approximations to the original kernel \ref{eqa:bowkernel}.
\begin{itemize}
\vspace*{-1ex}
\item {\bf Sensing 0 kernel:} $K^{0}(\mathbf{x}^{i}, \mathbf{x}^{j})
  := \log (K(\mathbf{x}^{i}, \mathbf{x}^{j}))$.
\vspace*{-1ex}
\item {\bf Sensing 1 kernel:} Following \eqref{eqa:bowkernel},
\vspace*{-1ex}
\begin{equation*}
K^{1}_{n} (\mathbf{x}^{i}, \mathbf{x}^{j}) :=\log \Biggl\{ \prod\limits_{w=1}^{W}
\frac{\Gamma( n \bar{x}_{w}^{i}+n\bar{x}_{w}^{j}+1)}{\Gamma ( n
  \bar{x}_{w}^{i} +1) \Gamma( n \bar{x}_{w}^{j} +1) } \Biggr\}
\end{equation*}
\vglue -2ex
\noindent 
where $\bar{\mathbf{x}}^{i} := \mathbf{x}^{i}/ N_i$ is the
word-frequency (normalized word-count) and $n$ is set to be some constant.
%
%
\vspace*{-1ex}
\item {\bf Sensing 2 kernel:} $K^{2}_N(\mathbf{x}^{i}, \mathbf{x}^{j})
  = \log ( K(\hat{\mathbf{x}}^{i}_N, \hat{\mathbf{x}}^{j}_N))$ where
  $\hat{\mathbf{x}}^{i}_N$ are the words counts obtained by randomly
  resampling the $i$-th document, uniformly, $N$ times.
\end{itemize}
In general, these approximations balance the number of words of
documents and avoid numerical issues by a log-mapping that compresses
the dynamic range.
%
In typical experimental settings, we have found that these
approximations
yield similar performance.
\vspace*{-2ex}
\section{Document Classification}
\label{sec:exp-text}
\vspace*{-1ex} 
In this section, we consider the problem of text document
classification. In this application context, the terms ``word'',
``vocabulary'', and ``document'' have their natural meanings.
We selected the 20 {\it Newsgourps}
dataset
which
contains 18774 news postings from 20 newsgroups (classes) to test
performance. The average number of words per document in this dataset
is 117. Following standard practice \cite{stopword:ref}, we removed a
standard list of stop words from the vocabulary.
\vspace*{-2ex}
\subsection{Binary Classification}
\label{sec:binary-text}
\vspace*{-1ex}
We chose to distinguish postings in the newsgroup {\it alt.atheism}
from {\it talk.religion.misc}. This is a difficult task due to the
similarity of content in these two groups \cite{DiscLDA08:ref,
  Zhu13:ref}.
The training set contains 856 documents and the test set contains 569
documents. The average number of words per document is 132.
\vglue -1ex
\begin{table}[hbt]
\vspace*{-0.1in}
\caption{Binary classification accuracy for {\it alt.atheism} vs. {\it
    talk.religion.misc} in the 20 Newsgroups dataset.}
\label{table:binarydoc}
\centering
\begin{tabular}{|c|c|c|c|}
\hline 
Method & CCR \% & Method & CCR \% \\ 
\hline 
Sensing1 SVM & 82.07 & RBF SVM& 78.91   \\ 
\hline 
Sensing2 SVM & 83.35 & DiscLDA \cite{DiscLDA08:ref} & 83.00\\ 
\hline 
PPK SVM\cite{Jebara04:ref}  & 81.02 & G-MedLDA \cite{Zhu13:ref} & 83.57 \\ 
\hline 
Diff SVM \cite{Lafferty05:ref} & 79.96&  &  \\ 
\hline 
\end{tabular} 
\end{table}
\vspace*{-2ex}
We used LIBSVM \cite{LibSVM:ref} to train our kernel SVMs. We report
results only for the Sensing 1 and Sensing 2 approximations of our
sensing-aware kernel since in this dataset, the number of words $N_i$
varies significantly across documents. Table~\ref{table:binarydoc}
compares the Correct Classification Rate (CCR \%) of the proposed
sensing-aware kernels against two model-based kernels, PPK and Diff,
and the baseline RBF kernel.
Table~\ref{table:binarydoc} also shows results for a discriminative
Latent Dirichlet Allocation method (denoted by DiscLDA)
\cite{DiscLDA08:ref} and a recent method that is based on a max-margin
supervised topic model (denoted by G-MedLDA) \cite{Zhu13:ref}. Both of
these methods posit more complex models and represent the current
state-of-the-art.
The free model parameters such as $n$ and $N$ in sensing kernels 1 and
2, $t$ in Diff, and $\sigma$ in the RBF kernel, were tuned using a 5
fold cross-validation. The CCR for the Gibbs MedLDA method is for the
best number of topics $k$ as in \cite{Zhu13:ref}. Words in the test
set that were not in the training set were dropped.

%
We can see that the proposed Sensing 1 and 2 kernels outperform the
RBF kernel and the model-based kernels PPK and Diff. Surprisingly, the
performance of our very simple method is better than that of DiscLDA
and is almost the same as that of G-MedLDA (with only 1 less document
correctly classified). These state-of-the-art methods make use of
fairly complex models and require sophisticated inference algorithms
while our sensing-aware kernel SVM makes minimal modeling assumptions
and has much lower time complexity.
%
%

%
\vspace*{-2ex}
\subsection{Multi-class Classification}
\vspace*{-1ex}
We next studied multi-class classification in the 20 Newsgroups
dataset with all 20 classes. We adopted a widely used training/test
split where the training set consists of 11,269 documents, and the
test set consists of 7,505 documents. We used the ``one-versus-all''
strategy following \cite{Zhu13:ref} to do multi-class classification
with binary classifiers. We followed the same settings as in the
binary case.  For the Sensing 1 and 2 kernels, we used $n = N = 150$.
Table \ref{table:multidoc} shows the CCRs for the proposed
sensing-aware kernels, the RBF baseline, the two model-based kernels
PPK and Diff, and the G-MedLDA \cite{Zhu13:ref} algorithm. Also shown
is the CCR for a recently developed deep learning method based on a
Restricted Boltzmann Machine (RBM) \cite{RBM12:ref} which outperforms
the standard RBF SVM. Deep learning has recently emerged as a powerful
generic approach and has attained state-of-the-art performance in many
applications. Since \cite{RBM12:ref} uses the same training settings,
we simply quote the results reported in that paper.
\vspace*{-1ex}
\begin{table}[hbt]
\vspace*{-0.1in}
\caption{Multi-class classification accuracy for the 20 {\it
    Newsgroups} dataset.}
\label{table:multidoc}
\centering
\begin{tabular}{|c|c|c|c|}
\hline 
Method & CCR \% & Method & CCR \% \\ 
\hline 
Sensing1 SVM & 79.5 & RBF SVM& 75.9 \\ 
\hline 
Sensing2 SVM & 80.5 & RBM \cite{RBM12:ref}& 76.2 \\ 
\hline 
PPK SVM\cite{Jebara04:ref}  & 78.2 & G-MedLDA \cite{Zhu13:ref} & 80.9   \\ 
\hline 
 Diff SVM \cite{Lafferty05:ref} & 78.0 & &\\ 
\hline 
\end{tabular} 
\end{table}
\vspace*{-2ex}
Observe that the approximate Sensing kernels 1 and 2 outperform all
other kernel SVMs and the RBM. As in binary classification, their
performance is close to the G-MedLDA method that uses complex models.
\vspace*{-1ex}
\section{Image Classification}
\vspace*{-2ex}
\label{sec:exp-image}
In this section, we consider the problem of recognizing image scene
categories.
We use the Natural Scene category dataset first introduced in
\cite{Li05:ref,Lazebnik06:ref}. This dataset consists of $15$ scene
categories, e.g., office, street view, forests, etc., with 200-400
grayscale images in each category (total 4485 images) and an average
image size of $300\times250$ pixels.
\vspace*{-2ex}
\subsection{Modeling Images as Bags of Features}
\vspace*{-1ex} There is extensive literature on BOF models for
images. A typical model consists of $1)$ local feature extraction,
$2)$ visual-vocabulary construction, and $3)$ image representation as
BoFs.
In order to highlight the effect of kernels, we follow the baseline
approach proposed in \cite{Li05:ref, Lazebnik06:ref, Grauman:ref}.

\noindent $1)$ {\it Local feature extraction}. For each image, the
SIFT descriptors of all the $16 \times 16$ image patches centered at
grid points spaced $8$ pixels apart are computed.

\noindent $2)$ {\it Visual vocabulary construction}. $D$ patches are
randomly selected from training images. $W$-means clustering is
performed over the corresponding $D$ SIFT descriptors. The $W$ mean
vectors form a visual vocabulary, labeled as $1,\ldots,W$.

\noindent $3)$ {\it BoF representation}. For each image, the SIFT
descriptor of each patch is assigned the label of its nearest neighbor
in the visual vocabulary. An image is then a collection of visual
words from the vocabulary and the spatial correlation is ignored.

To incorporate spatial structure, we use the pyramid matching scheme
proposed in \cite{Lazebnik06:ref}. This uses a sequence of
coarse-to-fine grids. At level $L$, an image is split into $2^L \times
2^L$ non-overlapping cells, and each cell is treated as a separate
``document''. Hence, at level $L$, each image $\mathbf{x}^{i}$ is
represented as a collection of $2^{2L} $ word-frequency vectors
$\{\mathbf{x}^{i}_{L,c}, c=1,\ldots,2^{2L}\}$, ordered according to
the spatial position of the cell. The kernel at level $L$ for two
images $\mathbf{x}^{i},\mathbf{x}^{j}$ is defined as
$K^{L}(\mathbf{x}^{i},\mathbf{x}^{j})= \sum_{c=1}^{2^{2L}}
\kappa(\mathbf{x}^{i}_{L,c}, \mathbf{x}^{j}_{L,c})$, where $\kappa$ is
the kernel function for two BoF documents.

The above kernel is only for a single level. The pyramid kernel $K$
for two images is defined as a weighted sum of single-level kernels:
$K(\mathbf{x}^{i},\mathbf{x}^{j}) = \sum_{l=0}^{L} \alpha^{l}
K^{l}(\mathbf{x}^{i},\mathbf{x}^{j})$. We use the same weighting
factors $\alpha^l$ as in \cite{Lazebnik06:ref}.
\vspace*{-2ex}
\subsection{Experimental Results}
\vspace*{-1ex}
We follow the settings in \cite{Lazebnik06:ref, Li05:ref}. We set the
vocabulary size as $W=400$.  
We randomly select 100 images per class for training and leave the
rest for testing.
We repeat experiments 10 times with a randomly selected
training/testing split for each run.
We use the ``one-versus-all'' strategy for multi-class classification
and report the average CCR over 10 random runs.

Following \cite{Lazebnik06:ref}, we consider three setups. ``$L=0$,
single'' uses $K^{0}$ as the kernel for classification. This simply
views the entire image as a document. Similarly, ``$L=2$, single''
uses $K^{2}$ as the kernel for classification. In this case, the image
is divided into $2^{2L} =16$ regions. ``$L=2$, Pyramid'' uses $K =
\sum_{l=0}^{2}\alpha^l K^{l}$ as the kernel value.

We compare our Sensing 0 and Sensing 2 kernels
with the probabilistic model-based PPK and Diff kernels and the
standard RBF kernel. We also compare with the classic Spatial Pyramid
(SP) kernel of \cite{Lazebnik06:ref}. This kernel is specially crafted
for this problem and has good empirical performance in several
Computer Vision tasks.  We also consider a recent work called
Reconfiguration BoW (RBoW), which makes additional modeling
assumptions on $p(\mathbf{z}|y)$.  Finally, we consider a recently
developed deep learning algorithm \cite{HiltonImg:ref} (denoted by
MRF) which achieves state-of-the-art performance on the same task.

For the Sensing 2 kernel, we simply set $N=500$. As in text
classification, all the other free parameters are tuned by a 5-fold
cross-validation.
We measure performance using the overall classification accuracy (CCR
\%). The results are summarized in Table \ref{table:multiscene}. Since
the settings are identical, we simply quote the results for MRF and
RBoW as reported in \cite{HiltonImg:ref} and \cite{RBoW12:ref}
respectively.
\vspace*{-1ex}
\begin{table}[!htbp]
  \centering
  \vspace*{-0.1in}
\caption{CCR \% for the scene category database for the three
  different settings}
\label{table:multiscene}
    \begin{tabular}{cccc}
    \toprule
          & $L=0$,single   & L=2, single   & L=2,Pyramid \\
    \midrule
    Sensing 0 & 72.3  & 79.1  & 80.7 \\
    Sensing 2 & 72.9  & 79.7  &  81.3 \\
    PPK \cite{Jebara04:ref}   & 73.6  & 77.7  & 78.6 \\
    Diff \cite{Lafferty05:ref}  & 74.7  & 77.1  & 77.8 \\
    RBF   & 65.6  & 72.2  &  73.5\\
    SP \cite{Lazebnik06:ref}    & 74.8  & 79.3  & 81.4 \\
    RBOW \cite{RBoW12:ref}  & N/A   & 78.6  & N/A \\
    \midrule
    MRF \cite{HiltonImg:ref}   & \multicolumn{3}{c}{81.2} \\
    \bottomrule
    \end{tabular}%
  \label{tab:addlabel}%
\end{table}%

We can see that in the ``$L=2$, single-level'' setting, the proposed
Sensing kernel 2 clearly improves over the classic kernel SP and
outperforms other kernels (RBF, PPK and Diff) as well as the more
complex RBoW method.
However in the ``$L=0$ single-level'' setting, the sensing-aware
kernels do perform worse than PPK, Diff, and SP.
This could be because $L = 0$ corresponds to the whole image. It has
been strongly suggested in the Computer Vision literature that
modeling the entire image as a BoF is unrealistic in many datasets
\cite{Grauman:ref} \cite{Lazebnik06:ref}.
Our proposed sensing-aware approach only makes sense if the data is
actually generated from the sensing structure. Our experimental
results accord with the fact that BoF is a reasonable model for local
regions, but not for the entire image.

We also did experiments for the ``$L=2$, pyramid'' setting. We can see
that the Sensing 2 kernel achieves almost the same performance as the
state-of-the-art SP and MRF.  The improvement is not as significant as
in the ``$L=2$ single-level'' case. This could be because it involves
the $L=0$ level kernel values, which does not fit the BoF model well.

\vspace*{-2ex}
\section{Conclusion}
\vspace*{-2ex}
In this paper, we proposed a novel kernel design principle to
incorporate partial model information.
On two distinct types of data, we showed that with minimal modeling
assumptions, the sensing-aware kernel improves upon other standard
kernels and handcrafted kernels for specific domains. We also observed
that the sensing-aware kernel can match the performance of the
state-of-the-art approaches.

\bibliographystyle{IEEEbib}
\bibliography{refs}

\end{document}